\documentclass[runningheads]{llncs}
\usepackage{graphicx}
\begin{document}
\title{Industrial Memories: Exploring the Findings of Government Inquiries with Neural Word Embedding and Machine Learning}
%
\titlerunning{Industrial Memories}

\author{Susan Leavy\and
Emilie Pine\and
Mark T Keane}
\authorrunning{S. Leavy et al.}
%
\institute{University College Dublin, Dublin, Ireland\\
\email{\{susan.leavy,emilie.pine, 
mark.keane\}@ucd.ie}}
%
\maketitle              
\begin{abstract}

We present a text mining system to support the exploration of large volumes of text detailing the findings of government inquiries. Despite their historical significance and potential societal impact,  key findings of inquiries are often hidden within lengthy documents and remain inaccessible to the general public. We transform the findings of the Irish government's inquiry into industrial schools and through the use of word embedding, text classification and visualization, present an interactive web-based platform that enables the exploration of the text to uncover new historical insights.



\end{abstract}
\section{Introduction}

The Irish Government published one of the most important documents in the history of the state in 2009. It spanned  over 1 million words and detailed the findings of a 9 year investigation into abuse and neglect in Irish industrial schools. The document however,  remains largely unread. The structure and style of the report obscures significant information and prevents the kind of system-wide analysis of the industrial school regime which is crucial to identifying recurring patterns and preventing the re-occurrence of such atrocities.

The Industrial Memories project, publicly available online\footnote{https://industrialmemories.ucd.ie}, presents a web-based interactive platform where the findings of CICA, commonly known as the Ryan Report, may be explored by researchers and the general public in new ways. The machine learning approach developed as part of this project addressed the challenge of learning within the context of limited volumes of training data. Techniques such as named entity recognition, word embedding and social network analysis are also used to extract information and unearth new insights from the report.


\section{System Overview}

\begin{figure}
\centering
\includegraphics[width=.7\textwidth]{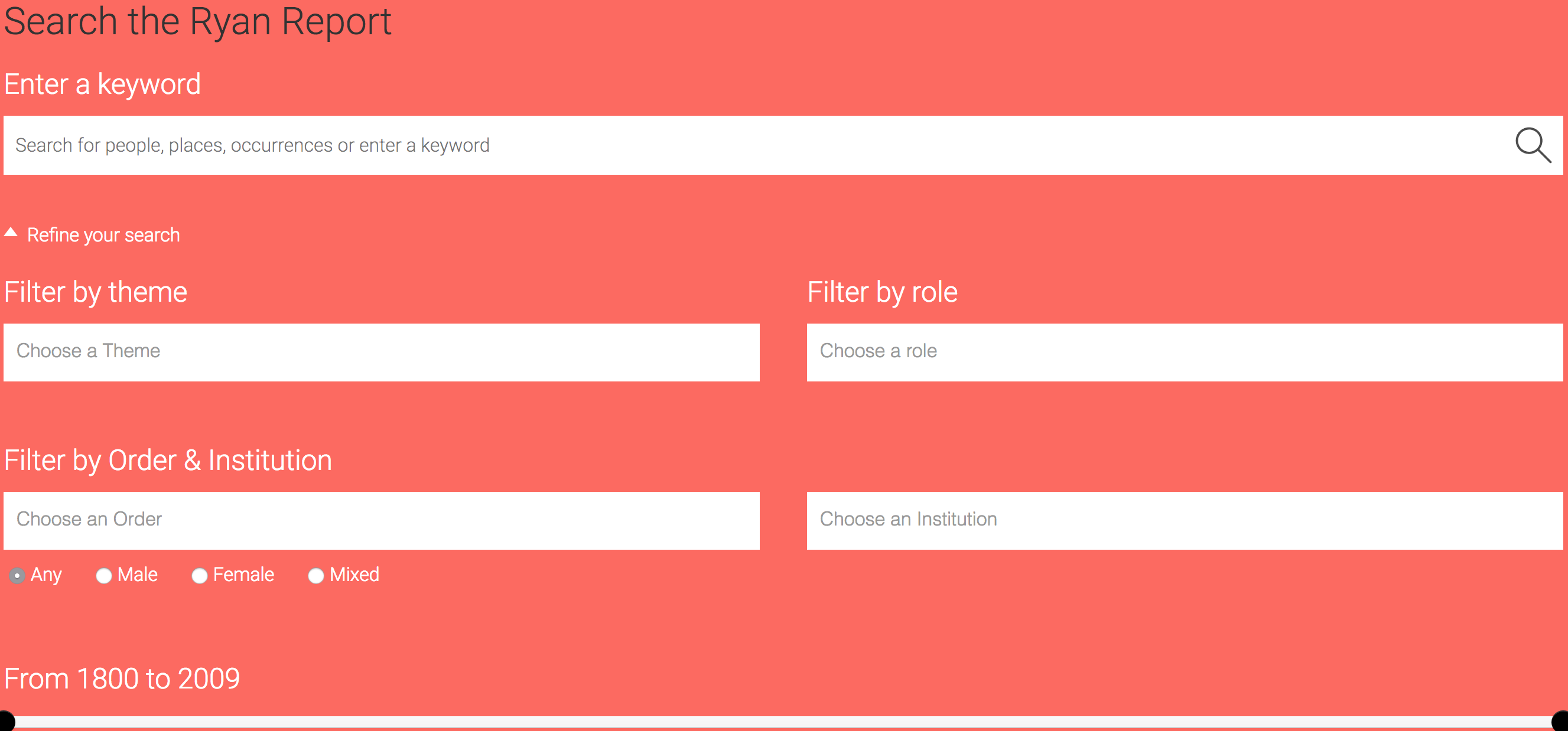}
\caption{Text Navigation Tool} \label{search}
\end{figure}

The Industrial Memories project developed a web-based platform where the narrative form of the Ryan Report is deconstructed and key information extracted. In the Ryan Report most of the chapters deal separately with individual industrial schools. This segregation of information prevents a system-wide analysis of common trends and patterns across the industrial school system. To address this, a random forest classifier was employed to annotate the report. An interactive web based platform enables users to extract information and interact with the report using a comprehensive search tool (Fig. \ref{search}).  Word embeddings were also used to identify actions pertaining to meetings and communication that occurred within the industrial school system facilitating an analysis of the dynamics of power and influence using social network analysis. Each individual named in the report was extracted using the Stanford NE system~\cite{finkel2005incorporating} which for the first time, allows the history of individuals to be traced over the entire industrial school system thus responding to a call for increased open data in government~\cite{kitchin2014data}.

\subsection{Semantic Classification of Text}

While some  semantic categories or themes were conducive to extraction using a rule based approach, categories including the transfer events, descriptions of abuse and witness testimony were extracted using a random forest learning algorithm. The random forest algorithm was chosen because, as an ensemble learner that creates a  `forest' of decision trees by randomly sampling from the training set, it is suited to learning from smaller datasets~\cite{polikar2006ensemble}. The Ryan Report was represented in a relational database with annotated excerpts comprising 6,839 paragraphs (597,651 words).  Each paragraph was considered a unit-of-analysis in the Report  for the purposes of annotation.
	


\subsection{Feature Extraction with Word Embedding}

Feature selection was based on the compilation of domain-specific lexicons in order to address the constraint in this project concerning low volumes of available training data. As with many projects that are based in the humanities, compiling training data is a manual process involving close reading of texts. To address this, domain-specific lexicons were compiled based on the entire dataset yielding high accuracy with low numbers of training examples.

Lexicons were generated using the word2vec algorithm from a number of seed terms~\cite{mikolov2013distributed}. Given that the word2vec algorithm performs best with larger datasets than the Ryan Report, five ensembles were generated and the top thirty synonyms for each seed term were extracted. Terms common across each ensemble were identified to generate domain specific synonyms. Features for the classification process were then derived from these lexicons.


 \begin{figure}
   \centering
\includegraphics[width=.7\textwidth,height=7.6cm]{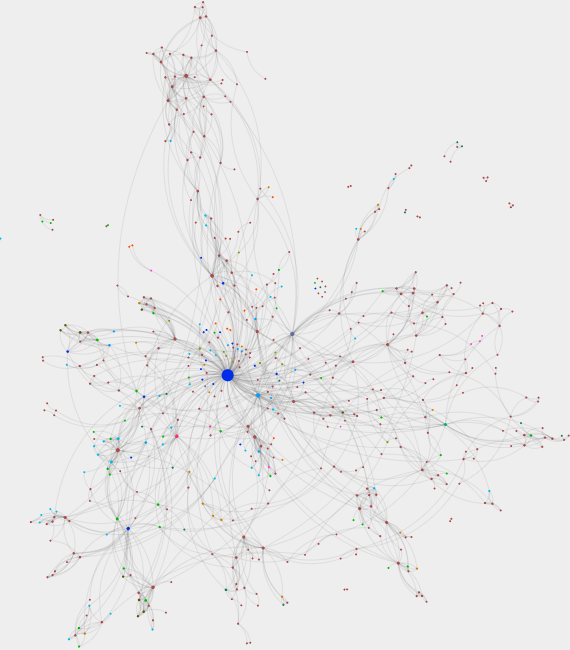}
\caption{Collocation Network of Entities in the Ryan Report} \label{maps}
\end{figure}

\subsection{Exploring Power, Influence and Migration within the System with Word Embedding and Network Analysis}

\begin{figure}
\centering
\includegraphics[width=.9\textwidth]{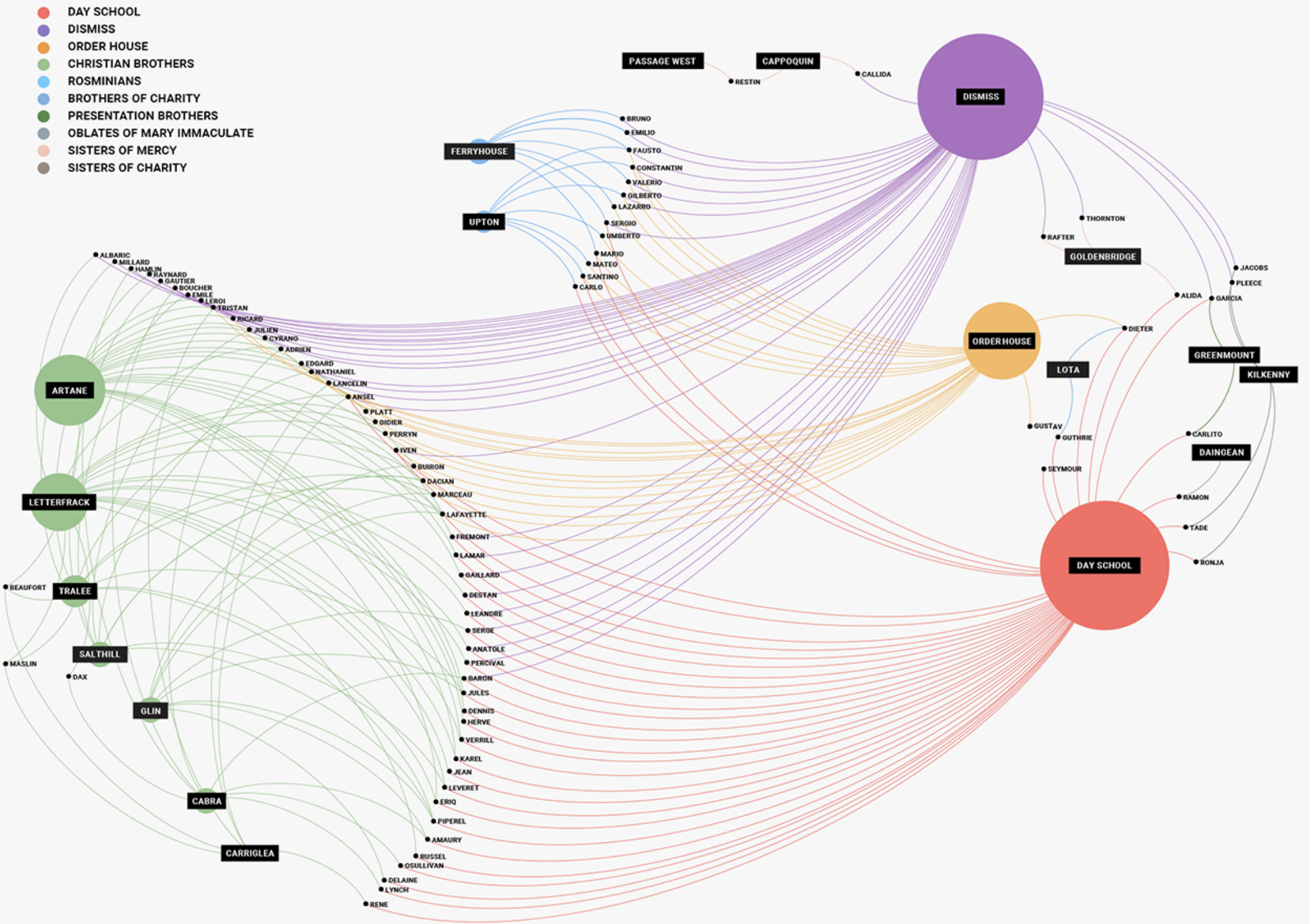}
\caption{System-Wide Analysis of the Transfer of Religious Throughout the Industrial School System} \label{transmap}
\end{figure}

Neural network analysis was used to generate lexicons describing interactions in the form of meetings, letters and telephone calls to be compiled and excerpts describing these communications extracted from the report. A social network based on these excerpts combined with entities named in the report was constructed along with a network detailing how entities were related in the narrative of the Ryan Report through collocation within paragraphs (Fig. \ref{maps}). The sub-corpus that was generated by automatically extracting excerpts detailing transfers was examined using association rule analysis revealing the recurring patterns governing the movement of religious staff often in response to allegations of abuse (Fig. ~\ref{transmap}). 

\section{Conclusion}

The Industrial Memories project presents a state-of-the-art text mining application  to navigate the findings of the Irish Government's Inquiry into Child Abuse at Industrial Schools. A demonstration video of the system is available online \footnote{https://youtu.be/cV1xzuJ0dv0}. Text analytic techniques were used to uncover new insights and system-wide patterns that are represented in an interactive web based platform accessible to the general public.

%
%

\bibliographystyle{splncs04}
\bibliography{mybibliography}

\end{document}